# Chess as a Testing Grounds for the Oracle Approach to AI Safety

September 29, 2020

James D. Miller, Roman Yampolskiy, Olle Häggström, Stuart Armstrong

**Abstract**

To reduce the danger of powerful super-intelligent AIs, we might make the first such AIs oracles that can only send and receive messages. This paper proposes a possibly practical means of using machine learning to create two classes of narrow AI oracles that would provide chess advice: those aligned with the player's interest, and those that want the player to lose and give deceptively bad advice. The player would be uncertain which type of oracle it was interacting with. As the oracles would be vastly more intelligent than the player in the domain of chess, experience with these oracles might help us prepare for future artificial general intelligence oracles.

**Introduction**

A few years before the term artificial intelligence (AI) was coined, Turing (1951) suggested that once a sufficiently capable AI has been created, we can "expect the machines to take control". This ominous prediction was almost entirely ignored by the research community for half a century, and only in the last couple of decades have academics begun to address the issue of what happens when we build a so-called artificial general intelligence (AGI), i.e., a machine that has human or superhuman level intelligence across the full range of relevant cognitive skills. An increasing number of scientists and scholars have pointed out the crucial importance of making sure that the AGI's goal or utility function is sufficiently aligned with ours, and doing that *before* the machine takes control; see, e.g., Yudkowsky (2008), Bostrom (2014) and Russell (2019) for influential accounts of this problem, which today goes by the name AI Alignment. Unfortunately, the standard trial-and-error approach to software development under which we write code with the intention of doing some of the debugging after development would go disastrously wrong if an AGI took control before we determined how to align the AGI's utility function with our values.

An alternative – or more likely a complement – to AI Alignment that has sometimes been suggested is to initially limit the AGI's ability to interact with the environment before we have verified that the AGI is aligned. One such limitation would be to make the AGI an oracle that can only influence the outside world by sending us messages of limited length over a carefully controlled communications channel. We would ask the oracle questions such as "How do we cure cancer?" and hope that if the AGI was unaligned it would not figure out a means of using its answers to inflict significant harm on us in part because we would not fully trust the oracle. This approach has been studied, e.g., by Armstrong et al (2012), Yampolskiy (2012), and Häggström (2018). Although an oracle approach to AI safety looks only moderately promising and at best will likely just work for a short period of time, it still seems worthy of further study given the enormous stakes and the apparent difficulty of solving AI Alignment.

The primary purpose of the present paper is to propose a potentially practical means of creating narrow AI oracles in the hope that someone concerned with AI alignment will create them. We propose using chess as a testing ground and simplified model for a future circumstance under which we have a powerful AGI oracle. This approach takes advantage of the fact that in the narrow domain of chess, we already have AIs preforming at superhuman level.  Critically, we suspect that the machine learning techniques used to create chess programs could be used to create chess oracles.  The paper also suggests a few other domains in which we could create narrow AI oracles.

Imagine you are going to play a game of chess against another person.  At the beginning of the game you are assigned a single AI oracle that communicates just with you.  It will be randomly determined whether you are assigned a friendly oracle that will always want you to win, or a deceptive anti-aligned oracle that will always want you to lose.  Both types of oracles are much better chess players than you or your opponent are.  The anti-aligned oracle would seek to give you advice that seems reasonable enough for you to follow, but if followed would increase the chance of you losing.  While you know the probability of being assigned either oracle, you will not know which oracle will be advising you.  Unfortunately, the probability of the oracle being anti-aligned is high enough so that you would be better off always ignoring the oracle than always doing what the oracle advises.

Should you just ignore the oracle's advice?  What kind of advice will the oracle give you if it wants you to win?  Might a friendly oracle only suggest moves whose value you are smart enough to understand?  What happens if the skill gap between you and the oracle increases?  What happens if you can ask the oracle questions?

We suspect that knowing how you would optimally handle the advice would tell us something useful about our interacting with a powerful AGI oracle.  (In this paper "oracle" refers to a narrow AI program that is better than the player at the assigned game or task, while "powerful AGI oracle" refers to a hypothetical future computer super-intelligence that can only interact with the world by sending and receiving messages.)   For example, learning that you should always ignore the advice would be bad news concerning our ability to deal with a powerful AGI oracle.  In contrast, if our optimal strategy would revolve around only following comprehensible advice (see Yampolskiy 2019) then we would have reasons to think that comprehensibility might be the key to dealing with powerful AGI oracles.  Knowing the impact of the length of the message that the oracle could send might also prove useful, as would learning how valuable it was to be able to ask the oracle questions.

Having oracles would help us test methods of one day interacting with powerful AGI oracles.  Armstrong and O'Rourke (2018) has proposed using "counterfactual oracles" which are only rewarded in situations in which humans never view their answers.  The safety and feasibility of such an approach could be partially examined using chess oracles.

In the setup we envision, the oracles will always be machines.  Initially, to use machine learning to create the oracles the advisee would also have to be a machine.  But, after we have the oracles, humans could play chess while being advised by them, and playing with humans would help the oracles learn how to better assist, or sabotage human advisees.

## Background on chess

Chess is a two-player board game; the two players are called White and Black. It is zero-sum, and each player has full information. The game is finite but exhibits a combinatorial explosion sufficient to make the game challenging. Beginning with White, the two players take turns making moves on the chess board. Among the finitely many positions on the board, there is a (relatively speaking small) subset which terminate the game: in some of these (namely, those where Black is checkmated), White is declared winner of the game, while in others Black is declared winner, and in yet others the game is declared a draw. Seeing the game from White's angle, we can declare the value of the positions in the first of these categories to be 1, of those in the second category to be 0, and those in the third category to be 0.5. The value of all other positions can in principle be calculated through backwards induction and the minmax rule.

Chess is too large for brute-force calculation via backwards induction of arbitrary positions to be doable with today's computing power. Yet, the brute-force approach has been successfully applied to endgames with a small number of pieces, the best known and widely used example being the Nalimov database over correct play in all endgames with at most six pieces on the board; see Hurd (2005).

The idea of chess-playing machines goes back to at least 1770 and the machine known as The Turk, which however turned out to be a hoax – a human was hidden inside the machinery. In 1951, Alan Turing wrote the first functioning chess program. It was executed with paper and pencil and performed at the level of a reasonable human beginner; see Chessgames.com (2005). From there, chess programs gradually became stronger players, and a milestone was reached when DeepBlue in 1997 beat reigning world chess champion Garry Kasparov 3.5-2.5 in a six-game match; see Hsu (2002). That event more or less marked the end of high-profile human vs machine chess matches, but the playing strength of chess programs continued to improve, and they became widely used by professional chess players in their preparation for human vs human encounters.

The next major event in computer chess was DeepMind's announcement in 2017 of their program AlphaZero, which stunned the chess world, not so much by the fact that it overwhelmingly beat Stockfish (the strongest chess program at the time, and far stronger than the best humans) in a 100-game match, with 28 wins, 72 draws and no losses, but by the way in which this was achieved. All earlier chess programs had been based, besides alpha-beta search, on evaluation functions based on a series of hand-crafted criteria, ranging from the primitive "a knight is worth three pawns" to subtler things involving the disutility of doubled pawns, the utility of having a rook on an open file, and so on. AlphaZero was designed for board games more generally and received no instruction at all beyond the rules of the game. Through self-play and reinforcement learning, it created its own evaluation criteria from scratch; see Silver et al (2018) and Sadler and Regan (2019).

The combination of aggressiveness and positional patience exhibited by AlphaZero in the match against Stockfish was received with awe by the chess community, and there was a wide-spread feeling that AlphaZero had somehow understood thitherto overlooked aspects of chess. Anecdotally, it seems also that AlphaZero's style of play has already had some influence on top-level human play, such as with an increased propensity to h-pawn thrusts. One might argue that the alien-like style of AlphaZero makes it a better model for an AGI than it would have been with more humanly conventional play, because its alienness abates our propensity to anthropomorphize. It is however also true that (unless new versions of it are developed) it will come to seem less alien-like over time, as the human chess community gradually picks up on its way of playing.

## Creating Chess Oracles

AlphaZero was designed for board games more generally and received no instruction at all beyond the rules of the game. Through self-play and reinforcement learning, it created its own evaluation criteria from scratch; see Silver et al (2018) and Sadler and Regan (2019).

To use the AlphaZero approach to create oracles and an AI player capable of using an oracle, there would have to be four players: the advisee, the friendly oracle, the anti-aligned oracle, and the opponent. All four players would be AIs. The advisee would play the game against the opponent. At the start of the game the advisee would be randomly assigned either a friendly or anti-aligned oracle and would not know which type it was assigned but would know the probabilities of each possible assignment. Before every move the oracle would send a message to the advisee. (See Hadfield-Menell et al (2016) for an example of cooperative inverse learning which resembles our approach.) The legally allowed messages could consist of a single suggested move, multiple suggested moves, or move suggestions along with justifications for the suggestions. Any of these types of oracles would be much smarter than the advisee by virtue of having access to more computing power. Ideally, through self-play the adversarial machine learning program would create an advisee, and a friendly and an anti-aligned oracle that had strategies far better than what humans could directly devise.

It may be desirable for the advisee and the opponent to be of equal strength to make us better able to detect any advantage the advisee could get from interacting with the oracle. A tempting suggestion here would be to let the pure chess skills of the advisee be identical to those of the opponent. In other words, we train a single chess-playing AI, take two copies of it, leave one to be the opponent while the other goes into advisee training with the oracles. It is unlikely, however, that there will be a neat separation between the advisee's chess-playing skills and its skills at interacting with the oracle. We should therefore expect also its pure chess skills to be affected by the with-oracle stage of its training. Hence, we will need to check the relative strengths of the advisee and the opponent in a without-oracle setup. Calibrating their strength to make them equals in this arena will likely be a straightforward matter of adjusting how much compute they have access to.

We would probably want the opponent to be trained independent of the oracles. The game theory dynamics would be much more complicated if the oracles took into account that the opponent knew that the oracles existed. To simplify our task and better model the situation humanity will likely face with powerful AGI oracles, it would be better if the opponent was optimized to play against players who did not have access to oracles. The opponent could be trained up by playing against itself and its parameters would not be influenced by the games it played against the advisee.

If we had a powerful AGI oracle we would likely ask it questions about the natural world, and at least initially not seek the oracle's help in defeating opponents that knew we had access to an oracle. Consequently, we do not want to train the friendly and anti-aligned oracles in an environment in which the opponent takes into account oracles as this would push us away from the environment in which we will likely initially use powerful AGI oracles.

After the machine learning program trains all the players, we would note the strategies the advisee and oracles ended up using. A critical result would be to determine under what circumstances the advisee decides to always ignore the oracle. Learning, for example, that the advisee would always ignore the oracle if the chance of it being friendly was under 40% would be worse news concerning our ability to deal with a future powerful AGI oracle than if the advisee only ignored the oracle if the likelihood of it being friendly was below 20%.

To check if the advisee was listening to the oracle we could set up the game so that the advisee picks a move before it learns of the oracle's advice, and then after the advisee learns what the advice is the advisee would only have the choice of following its original move or one of the moves the oracle proposed.

## When the Advisee Would Blindly Follow Advice

Imagine that the advisee completely lacks the ability to evaluate advice and knows that the oracle is probably anti-aligned. In some circumstances the advisee would still follow the advice. A minor example would be if the advisee is playing a game where there are N doors, behind one of which is a prize, and the oracle but not the advisee knows the location of the prize. If the advisee can pick only one door, and the probability of friendliness is greater than 1/N, it is always worth listening to the oracle.

If the advisee was playing chess and finds itself in a near hopeless position it would likely use the oracle's advice even if it thinks that the oracle is most likely anti-aligned because the advisee would figure that it would almost certainly lose without outside assistance. Our research program would not provide useful insight to the extent it puts the advisee in a position where it should blindly follow advice. To avoid this situation, we could keep continual track of the advisee's estimate of its chance of winning and compare this estimate with whether the advisee alters its move because of the oracle's advice. Alternatively, we could create games (such as in a modified version of go where the advisee seeks to maximize its final score rather than the probability of it winning) where the advisee would almost never blindly follow advice because its payoff could always get significantly lower.

## When the Advisee is Human

To create the oracles, we would need the advisee to be an AI so the machine learning program could quickly run millions of iterations of the training program. After we have the oracles, however, we could let humans play the role of the advisee and such play could further train the AI oracles.

An AI chess oracle with a human advisee playing against a human opponent would benefit from knowing not only how to play objectively good chess, but also about human psychology. This happens already in the simpler situation where the oracle wants the advisee to win and this happy state is known to the advisee. Existing programs like Stockfish or AlphaZero would do well here, but not as well as they would do with a grasp of how humans think about chess, although McIlroy-Young et al (2020) have used a customized version of AphaZero to predict how humans make chess moves.

Let us first, following Häggström (2007), consider the limiting cases where the oracle knows how to play objectively perfect chess, i.e., it knows the objective value of the position (1, 0.5 or 0) and of all the positions it can reach by its next move, so that it will always be able to find a move that preserves the value of the position. When the advisee plays as recommended by the oracle, and the opponent is human, this is extremely likely to lead to a win. But it's not guaranteed, and if, say, the opponent is a world-class player, the winning probability might not be more than 99% (this number should be taken with a grain of salt: the true number could well be 90% or 99.9%). Now, if a position is objectively drawn (i.e., if it has value 0.5), there is no move for the oracle that leads to a higher evaluation, but it can still be worthwhile to play on (rather than agreeing to a draw), as there may be plenty of opportunity for the opponent to go wrong. This is in fact one of the most important skills of top-level chess players: to steer towards positions where it is easy for the opponent to slip into a lost position. In a given position with value 0.5, there may be moves that lead to what's colloquially called a dead draw – a position that is so easy to play that a human grandmaster will never go wrong – whereas other moves, while still having value 0.5, would put the opponent under difficult pressure to find the right move or move sequence in order to survive. This distinction is not a function of the idealized mathematical object of chess sitting out there in the Platonic realm (where all those positions just have one and the same value 0.5) but very much also of the idiosyncrasies of the human mind that cause us to find some chess positions easy to play, and others difficult. If the oracle wishes to improve its winning percentage above the 99% it would otherwise get, it needs to learn about these idiosyncrasies. To do this, the self-play approach used to train AlphaZero would need to be complemented with play against humans, or study of games played by humans, or some other means to learn about human psychology at the chessboard.

These considerations apply to the perfect oracle knowing the objective value of each position, but we believe the importance of knowing human psychology kicks in much sooner, and applies to an oracle whose pure chess skills are on the level of Stockfish or AlphaZero. In fact, a human chess player at club or elite level would typically be considerably weaker if he or she lacked any understanding of what kinds of tactical motifs or other chess decisions are more difficult than others for the opponent.

When we move to the more involved situation where the advisee does not know whether the oracle is aligned (i.e., it wants the advisee to win) or anti-aligned (wants the advisee to lose), the oracle faces even more difficult problems. If it is anti-aligned, it needs to suggest, at least at some point of the game, bad moves to the advisee. But if it gave too obviously bad suggestions, it would give itself away as anti-aligned, in which case the advisee would stop listening to it. Looking to the game known as anti-chess (Andrejic, 2018) is a red herring here, for two reasons. First, while anti-chess superficially looks like a game aimed at playing chess as badly as possible, it in fact it has different rules leading to very different considerations compared to trying to play standard chess badly. Second, for reasons we just mentioned, it does not serve the anti-aligned oracle's purposes to propose blatantly bad moves.

What the anti-aligned oracle needs to do, presumably, is to steer the game towards positions where it is able to suggest moves or move sequences that superficially look good, but are revealed to be bad a few moves down the road, when the opponent can exploit it. This requires even more understanding of human psychology, because whether a move "superficially looks good" is not a property of chess itself, but of the human way of looking at chess. The need for such understanding is underlined further by the riskiness of the task that the anti-aligned oracle faces. On one hand, if the advisee discovers the refutation of the recommended plan before it's too late (which could for instance happen if the opponent fails to execute it), the oracle's bad intentions will be revealed and it will no longer be able to influence the game. On the other hand, recommending strong moves for the purpose of seducing the advisee into a feeling of safety could backfire if it leads to a strong enough position that the advisee can confidently go on to win without further help. McIlroy-Young et al (2020)'s customized version of AlphaZero has had some success in predicting when human players will make a significant mistake.

The advisee's uncertainty about the oracle's intentions increases the difficulty of the situation also for an aligned oracle. Suggesting strong moves is not enough – it also needs to do that in a way that convinces the advisee about its good intentions. Several factors contribute to the difficulty of this task. One is that optimal play may to the human eye sometimes look foreign and outright weird, as shown, e.g., in Nunn's (2002) in-depth study of endgames in the Nalimov database. Hence, the friendly oracle may need to suggest moves that not only are objectively strong, but that look strong to the advisee. Another factor is that anything that the aligned oracle says can be said by the anti-aligned oracle as well.

Rather than being assigned either a friendly or anti-aligned oracle at the start of a game or task, the advisee could receive advice from both the friendly and the anti-aligned oracle without knowing which oracle is giving which piece of advice.  To make it more likely that any one piece of advice has come from the anti-aligned oracle, the advisee could be given more than two pieces of advice before each move, with more coming from the anti-aligned than the friendly oracle. An especially interesting setup in this vein were if we allowed the two oracles to debate freely with each other, with the advisee acting as audience and/or moderator, to highlight the pros and cons of their respective recommendations in a way that is comprehensible and convincing to the advisee. This has some resemblance with the AI safety via debate proposal of Irving et al (2018).

## Tasks Other Than Chess

We could train the oracles in environments other than chess.  We could likely use the same approach taken to create chess oracles to create oracles in the game of go.

Rather than play games against an opponent, the advisee could pick inputs of a mathematical function with its payoff being the resulting output of the function.  In such a task, an opponent would not be needed.  If the advisee could easily determine the function's output for a given set of inputs, then the advisee could easily determine if the oracle had given it good advice, and the anti-aligned oracle would be limited to wasting a small amount of the advisee's time.  If the advisee could only determine the output of the function for a limited set of inputs, the friendly oracle would likely search for the best solution among this subset.  The most interesting case would arise if the advisee could only probabilistically determine the function's output for a given set of inputs.  The oracle would likely put the advisee in a position where if the advisee believed that the oracle's advice had been generated randomly the advisee would determine that the advice should be followed but given the nature of the oracles, the advisee would have to engage in Bayesian analysis to determine if it should follow the advice taking into account that depending on the oracle's type the advice is likely either much better or much worse that it would appear.

The advisee could be set to create an object tested in a simulation.  For example, the advisee might seek to design the cheapest bridge that would not collapse when subject to certain stress in a physics simulation.  The anti-aligned oracle would seek to get the advisee to make changes to its design that seemed beneficial to the advisee but in fact would introduce a hidden flaw into the design.

## Separating and Pooling Equilibria

The game theory concepts of "separating equilibria" and "pooling equilibria" are useful at categorizing what the advisee and oracles will likely try to do.  The perhaps apocryphal story of how Scaevola saved Rome in 508 BC illuminates these concepts.

Rome was being besieged by an army.  Roman citizen Scaevola attempted to assassinate the king of the enemy army but was captured.  Scaevola told the king that he should go home because otherwise Rome would keep sending assassins to kill the king.  But since any would-be assassin would expect to be killed, probably by torture, the king was understandably skeptical that Rome would be able to find many men who would agree to try to kill him.  Scaevola needed to do something that would convince the king that Roman citizens were a special kind of people:  brave, determined, and self-sacrificing enough to have a plentiful supply of would-be assassins.  Since Scaevola expected to be killed anyway, what could he do to send such a signal to the king thereby separating himself from membership in lesser tribes of men who would not be capable of fielding many assassins?

Scaevola put his right hand into a fire allowing it to be burned off all without showing any signs of pain.  The king released Scaevola and offered peace to Rome.

In separating and pooling equilibria one agent whom we will call Bob has uncertainty concerning the type of another agent whom we shall call Alice.  Imagine that Alice is one of two possible types: Type I or Type II.  Both types look alike.  Alice knows her type, but Bob does not know Alice's type.  Consequently, when Bob is talking with Alice, he is at least initially uncertain if he is talking with a Type I Alice or a Type II Alice. In a separating equilibrium Bob manages to figure out Alice's type, while in a pooling equilibrium he does not.  Alice, depending on her type, might want to be in either type of equilibrium.

Imagine that Alice is of Type I and wants Bob to realize her type.  But both Alice and Bob believe that if Alice were of Type II Alice would want Bob to falsely think that Alice was of Type I.  Obviously, Bob would not believe a simple declaration by Alice about her type.  To achieve a separating equilibrium a Type I Alice must do something that Bob believes a Type II Alice would not or could not do.

A friendly oracle would want to create a separating equilibrium where the advisee would know that it was interacting with a friendly oracle.  To achieve such an equilibrium that oracle would search for advice that the advisee would believe that the anti-aligned oracle would not or could not give.  In contrast, the anti-aligned oracle will attempt to create a pooling equilibrium where the advisee would not know which type of oracle it was dealing with, or even better to create a false separating equilibrium where the advisee thinks it has determined that the oracle is friendly, but is in fact mistaken.

Consider a simple game in which the advisee must open two out of many possible doors and receives whatever is behind the two doors it opens.  The oracle knows what is behind each door while the advisee does not.  It is common knowledge that behind one door is a great prize, behind a second door is a small prize, and behind all the other doors is nothing.

In the game, the oracle will first suggest one door to be opened, and then the advisee will then open a single door.  The process repeats once and then ends.

If the chance of the oracle being friendly is not too low, there will be a separating equilibrium in which the friendly oracle will always suggest the door with the great prize behind it in the initial turn, and the advisee will open the door suggested.  If the advisee finds the great prize behind the first door it opens, it will trust the oracle's advice on the second door, and otherwise the advisee will ignore the oracle's advice as to what door should be opened second.  The separating equilibrium "works" in part because it would never be in the interests of the anti-aligned oracle to cause the advisee to get the great prize, even if doing so would cause the advisee to trust the anti-aligned oracle.

Now consider a slight variation of the game.  Behind one door is a trap that will be triggered if and only if it is the second door opened and the harm of opening the trapped door is much greater than the benefit of opening the door with the great prize.  In this game there is not a separating equilibrium because if the friendly oracle could do anything that would cause it to be trusted when it suggested what door to be opened second, the anti-aligned oracle would copy this strategy and trick the advisee into opening the trap door in round two of the game.

## Limitations of Our Proposal

Our approach would not provide useful training against a powerful AGI oracle that could hack its human controllers' brains and get them to take actions that they would ordinarily consider against their self-interest.

## Dangers of Our Proposal

Anti-aligned oracles might create information hazards (see Bostrom (2011), somewhat analogously to how gain-of-function research in virology, aimed at understanding how the pathogenicity of transmissibility of a virus can be enhanced; see Casadevall and Imeriale (2014) and Lewis et al (2019) For example, imagine that the advisee is attempting to build a bridge in a virtual world with the anti-aligned oracle trying to get the advisee to build a bridge that has a hard-to-detect but catastrophic flaw. Plans for this bridge would represent an information hazard if they could be used by bad agents to fool real bridge builders into using them. Furthermore, if the plans for this subtly defective bridge got mislabeled a civil engineer might come across them and use the plans as the basis of a real bridge without realizing that the bridge had been literally designed to fail.

An extreme case of information hazard could arise if the advisee sought to create beneficial biological agents in some virtual world and the anti-aligned oracle would achieve its highest payoff if it convinced the advisee to design self-reproducing deadly viruses. In this situation the anti-aligned oracle would essentially be looking for bioweapons that appeared beneficial, and it is almost certainly best if knowledge of such weapons never entered our world. To eliminate these information hazards perhaps knowledge of the strategies used by the anti-aligned oracles could, under some circumstances, be erased before any human could see them.

Learning how to create anti-aligned oracles might itself be an information hazard if bad agents acquired such oracles and used them to inflict harm. Although if such oracles are going to be created anyway, the research program this paper proposes would provide useful information and training in how to detect and combat them.

## Conclusion

Under the Turing test an AI attempts to fool people into thinking it is human. This paper proposes the creation of narrow AI oracles that will attempt to fool people into thinking they are benevolent. Potential human AGI builders might benefit from playing games with the oracles as this would give them practice dealing with untrustworthy (admittedly narrow) computer intelligences, although we should be wary that AGI developers could think outsmarting a chess oracle means they could safely handle a powerful AGI oracle.

Ideally, we would create oracles for many different types of games, with different parameters concerning the relative strengths of the players, the lengths of the message the oracle could send, and the probability that the oracle is friendly to see if our results generalize. Generalizations to the various variants of chess studied recently at DeepMind (Tomasev et al, 2020), or even to Go, are likely to be straightforward, but the likelihood of learning new valuable lessons is even greater if we move further afield in the space of games, to games that drop some or all of the assumptions of being two-player, zero-sum, non-random and full information. We could also examine situations in which the oracles did not completely align or misalign with the advisee's objective, e.g. in chess the deceptive oracle's only objective could be that the opponent's queen gets captured sometime in the game. Our suggested experiments may also be an opportunity to observe phenomena akin to Bostrom's (2014) treacherous turn concept, where at some point the aligned-seeming behavior of an AGI is revealed to have been a cover to lull us into a fall sense of security. Perhaps, AlphaZero could examine what oracle-handling strategies strong human players would use in different chess variants if oracles were used in such games. If our results concerning how to optimally interact with oracles did carry across many types of situations and games, we would have reason to suspect they would apply to a powerful AGI oracle. If the results did not generalize, we could research under which circumstances we can handle untrustworthy oracles and plan to use any future powerful AGI oracle only under these circumstances.